\documentclass[conference]{IEEEtran}
\IEEEoverridecommandlockouts

\usepackage{cite}
\usepackage{amsmath,amssymb,amsfonts}
\usepackage{algorithm}
\usepackage{algpseudocode}
\usepackage{graphicx}
\usepackage{textcomp}
\usepackage{xcolor}
\usepackage{hyperref}  
\usepackage{microtype}  
\usepackage{times}
\usepackage{latexsym}
\usepackage{multirow}
\usepackage{float}
\usepackage{subfigure}
\usepackage{booktabs}

\def\BibTeX{{\rm B\kern-.05em{\sc i\kern-.025em b}\kern-.08em
    T\kern-.1667em\lower.7ex\hbox{E}\kern-.125emX}}
\begin{document}

\title{Hybrid Multi-stage Decoding for Few-shot NER with Entity-aware Contrastive Learning}

\author{
    \IEEEauthorblockN{Congying Liu$^{1}$\IEEEauthorrefmark{1}, Gaosheng Wang$^{1,2}$\IEEEauthorrefmark{1}\thanks{\IEEEauthorrefmark{1} The first two authors have equal contributions to this paper.}, Peipei Liu$^{1,2}$\IEEEauthorrefmark{2}\thanks{\IEEEauthorrefmark{2} Corresponding Author: peipliu@yeah.net}, Xingyuan Wei$^{1,2}$, Hongsong Zhu$^{1,2}$}
    \IEEEauthorblockA{$^{1}$University of Chinese Academy of Sciences, Beijing, China}
    \IEEEauthorblockA{$^{2}$Institute of Information Engineering, Chinese Academy of Sciences, Beijing, China}
}

\maketitle

\begin{abstract}
Few-shot named entity recognition can identify new types of named entities based on a few labeled examples. Previous methods employing token-level or span-level metric learning suffer from the computational burden and a large number of negative sample spans. In this paper, we propose the Hybrid \textbf{M}ulti-\textbf{s}tage Decoding for \textbf{F}ew-shot \textbf{NER} with Entity-aware Contrastive Learning (\textbf{MsFNER}), which splits the general NER into two stages: entity-span detection and entity classification. There are 3 processes for introducing \textbf{MsFNER}: training, finetuning, and inference. In the training process, we train and get the best entity-span detection model and the entity classification model separately on the source domain using meta-learning, where we create a contrastive learning module to enhance entity representations for entity classification. During finetuning, we finetune the both models on the support dataset of target domain. In the inference process, for the unlabeled query data, we first detect the entity-spans, then the entity-spans are jointly determined by the entity classification model and the \textit{KNN}.
We conduct experiments on the open FewNERD dataset and FewAPTER dataset, the results demonstrate the advance of \textbf{MsFNER}.
\end{abstract}

\begin{IEEEkeywords}
Few-shot learning, named entity recognition, meta-learning  
\end{IEEEkeywords}

\section{Introduction}
\label{sec:intro}

Named Entity Recognition (NER) is a fundamental task in Natural Language Processing (NLP), involving the identification and categorization of text spans into predefined classes such as people, organizations, and locations \cite{yadav2018nersurvey,lijing2022nersurvey}. 
Although traditional deep neural network architectures have achieved success in the fully supervised named entity recognition task with sufficient training data \cite{lample2016neuralner,mahovy2016neuralner}, they are difficult to adapt to the dynamic nature of real-world applications, which require the model to quickly adjust itself to new data or environmental changes.
In this case, researchers have proposed the few-shot named entity recognition (few-shot NER) to explore entity recognition with limited labeled data \cite{fritzler2019few}. Few-shot NER enables existing models to quickly transfer learned knowledge and adapt to new domains or entity classes.

Specifically, the few-shot NER model is first trained on the source domain dataset $D_{source} = (S_{source}, Q_{source})$ and then the trained model is transferred to new target domain dataset $D_{target}=(S_{target}, Q_{target})$ to infer for $Q_{target}$ \cite{Snell2017tokenfewner1}. 
$S_{target}$ represents the support dataset comprising $N$ entity types ($N$-way), and each type is exemplified by $K$ annotated examples ($K$-shot). The $Q_{target}$ is the query dataset with the same entity types as the support dataset. 
Few-shot paradigm can offer a flexible and cost-effective solution to the adaptability challenge, making it a focal point of research to enhance the performance of NER systems in scenarios with limited labeled data or emerging entity types.

Currently, there are two mainstream researches for few-shot NER: token-level \cite{fritzler2019few,hou2020fewshotnerthransferlearning,yangyi2020simplefewshotner,das2022container} and span-level metric learning methods \cite{wangpeiyi2022esd,yudian2021fewshotner,matingting2022mamlprotype}. 
In token-level methods, each token is assigned an entity label based on its distances from the prototypes of entity classes or the support tokens. However, these approaches often have high computational costs and fail to maintain the semantic integrity of entity tokens, leading to increased susceptibility to interference from non-entity markers. On the other hand, although span-level methods can mitigate the partial issues associated with token-level approaches by evaluating entities as spans, all possible spans are enumerated would result in an N-square complexity and an increase in noise from a large number of negative samples. 

Considering the challenges, we hope to solve the following problems: 1) To improve the few-shot NER identification efficiency, how can we encourage the semantic divergence between entities and non-entities to determine effective entity spans? 2) To improve the entity span classification, how can we control and coordinate the semantic distance of different entity types to make the semantic representations of entities within the same types be proximate while those in disparate types be distant?

In this paper, we propose a hybrid multi-stage decoding approach for few-shot NER with contrastive learning (i.e., \textbf{MsFNER}). 
Specifically, \textbf{MsFNER} splits the general NER into two stages and this paper describes them by three processes: training process, finetuning process, and inference process. The training process consists of two objectives: training the best entity span detection model and the best entity classification model separately based on the supervised support dataset of source domain. In this process, we regard the training task as a meta-learning task and train entity detection model by employing Model-Agnostic Meta-Learning(MAML)\cite{finn2017maml},
and train entity classification model through the entity-aware contrastive learning and 
MAML-ProtoNet\cite{matingting2022mamlprotype}.
In the finetuning process, we finetune the trained entity detection model and entity classification model on the support dataset of target domain with the same pattern as training process. 
In the inference process, we deploy the \textit{KNN} and two finetuned models on the query dataset of target domain to predict entity spans and their types, and there are four phases for prediction. In the first phase, we build a \textit{key-value} datastore $\mathcal{D}_{knn}$ on the support dataset $S_{target}$, where \textit{key} is the entity representation and \textit{value} is the corresponding label. Subsequently, we apply the entity detection model to get entity spans. With the entity detection model, we could reduce the impact of negative samples and reduce computational complexity. After that, we pass the representations of detected entity spans into the entity classification model to produce their predicted type distribution $p_{soft}$ while we also feed the representations into \textit{KNN} to obtain the predicted results $p_{knn}$ based on the $\mathcal{D}_{knn}$. In the final phase, we combine $p_{soft}$ and $p_{knn}$ for predicting entity types.

We summarize our contributions as follows:
\begin{itemize}
    \item We propose a hybrid multi-stage decoding for few-shot NER with entity-aware contrastive learning, in which we first detect entity spans to improve efficiency and then employ contrastive learning or \textit{KNN} to augment entity classification performance.
    \item Experimental results show that our model achieves new SOTA performance compared with previous works. We also evaluate the few-shot NER task on LLM ChatGPT, and 
    \textbf{MsFNER} can have the comparable comprehensive results with ChatGPT in terms of performance and efficiency. 
\end{itemize}

\section{Methodology}
\label{sec:method}

In this section, we will introduce the details of \textbf{MsFNER}. The following content is arranged according to the three stages of the model.

\begin{figure}[!ht]
  \centering
  \includegraphics[height=3.2in,width=3.4in]{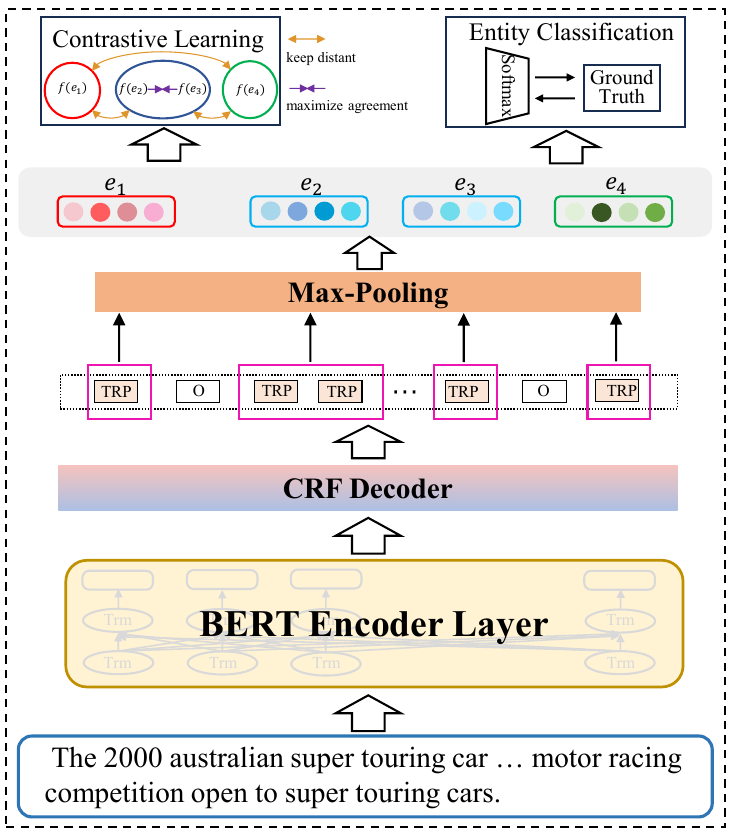}
  \caption{The schematic diagram of the training process.}  
  \label{trainingstage}
  \vspace{-0.4cm}
\end{figure}

\subsection{Training Process}
In this process, we introduce the separate process of training and finding the optimal entity span detection model and entity classification model.

\textbf{Entity Span Detection (ESD)}

\subsubsection{ESD Method}

In this module, we regard the entity span detection as a sequence labeling task with the \textit{BIOES} tagging scheme. The tag set $L$ = $\{$\textit{B, I, O, E, S}$\}$ means that we only care about the boundaries of the entities without entity types.

For a given sentence $x$ with $n$ tokens $x$=$(x_1,x_2,...,x_n)$ in $D_{source}$, we first encode it using the pre-trained language model BERT. After that, we can obtain the contextualized representations $h = (h_1, h_2, ..., h_n)$ for all tokens.

The sequence token representations are then sent into a \textit{CRF} to detect entity spans. 
Finally, the training loss in $D_{source}$ can be depicted as:
{
  \setlength{\abovedisplayskip}{1pt}
\begin{equation}
   L_{ESD} = -\sum log P(y|x)
\label{esdloss}
\end{equation}}
{\setlength{\abovedisplayskip}{-0.5pt}
  \setlength{\belowdisplayskip}{1pt}
\begin{equation}
    P(y|x) = \frac{\prod_{i=1}^{|x|} \varphi_i(y_{i-1}, y_i, x)}{\sum_{y^{'}}\prod_{i=1}^{|x|}\varphi_i (y_{i-1}^{'}, y_i^{'}, x)}
\label{crfprob}
\end{equation}}where $\varphi_i(y_{i-1}, y_i, x)$ and $\varphi_i (y_{i-1}^{'}, y_i^{'}, x)$ are potential functions, $y'$ is the possible label sequence.
\subsubsection{Training}
We regard the training process as a meta-learning task and divide the $D_{source}$ into many individual meta learning tasks. Then, we train the best entity-span detection model by MAML .

Specifically, the model first randomly samples a batch of task data $\mathcal{T} = \{(S_{source}^{(i)}, Q_{source}^{(i)})\}_{i=1}^{bz}$ from the source domain data $D_{source}$, where the batch size is $bz$. Next, the model is trained separately on $\mathcal{T}_s = \{S_{source}^{(i)}\}_{i=1}^{bz}$ and $\mathcal{T}_q = \{Q_{source}^{(i)}\}_{i=1}^{bz}$, which we refer to as inner-loop update (Task-level adaptation) and outer-loop update (Meta-optimization), respectively. An overview of the training process is presented in Algorithm \ref{algo:MAML4Span}.

In the inner-loop update, for each learning episode in meta-learning, the model performs one gradient update on the task data $S_{source}^{(i)}$ to update the parameters:
\begin{equation}
    \theta_{i}^{\prime}=\theta-\alpha \nabla_{\theta} {L}_{\mathit{ESD}}\left(S_{source}^{(i)}; {\theta}\right)
\label{onegradup}
\end{equation}
where $\theta = \{\theta^{encoder}, \theta^{crf}\}$ represents the model's learnable parameters, $\theta_{i}^{\prime}$ is the updated parameter after one gradient update on the task data $S_{source}^{(i)}$, and $\alpha$ is the learning rate used to minimize the loss function (Equation (\ref{esdloss})).

After the inner-loop update, the model will fully utilize the labeled query set $\mathcal{T}_q$ from the source domain for the outer-loop update. This update differs from the inner-loop update's model optimization approach (i.e., using the support set of one task in each learning episode for parameter optimization). Instead, it follows the conventional neural network training method, where the loss from all the training data in the batch is optimized together. The main purpose of using both inner and outer-loop update is to enhance the model's generalization ability on new task data and prevent overfitting to the support set data. The optimization objective of the outer-loop update is:
\begin{equation}
    \min _{\theta} \sum\nolimits_{\mathcal{T}_q} {L}_{\mathit{ESD}}\left(Q_{source}^{(i)};\theta_{i}^{\prime}\right)
\label{mamllosssum}
\end{equation}
Although the loss calculation at this point involves $\theta_{i}^{\prime}$, the model's goal remains to update and optimize $\theta$. The outer-loop update optimization is performed through Stochastic Gradient Descent (SGD), and therefore, the model parameter update process can be expressed as:
\begin{equation}
    \theta=\theta-\beta \sum\nolimits_{\mathcal{T}_q} \nabla_{\theta_i^{\prime}}{L}_{\mathit{ESD}}\left(Q_{source}^{(i)};\theta_{i}^{\prime}\right)
\label{mamlgradupdate}
\end{equation}
where $\beta$ is the learning rate for SGD.

\begin{algorithm}
\caption{Training Process of Entity Span Detection Model Based on MAML}
\begin{algorithmic}[1] 
\Require $D_{source}$: Training dataset for the source entity types
\Require $\alpha, \beta$: Learning rates for inner and outer-loop update
\State Randomly initialize $\theta$
\While{not done}
    \State Randomly select a batch of task data $\mathcal{T}$, where $\mathcal{T} \sim D_{source}$
    \ForAll{$\mathcal{T}$}
        \State Select one task data $(S_{source}^{(i)}, Q_{source}^{(i)})$ from $\mathcal{T}$
        \State Compute the gradient $\nabla_{\theta} {L}_{\mathit{ESD}}\left(S_{source}^{(i)}; {\theta}\right)$ based on the support set $S_{source}^{(i)}$
        \State Perform a gradient update: $\theta_{i}^{\prime}=\theta-\alpha \nabla_{\theta} {L}_{\mathit{ESD}}\left(S_{source}^{(i)}; {\theta}\right)$
        \State Compute the gradient $\nabla_{\theta_i^{\prime}}{L}_{\mathit{ESD}}\left(Q_{source}^{(i)};\theta_{i}^{\prime}\right)$ based on the query set $Q_{source}^{(i)}$, and store the result
    \EndFor
    \State Outer-loop update: $\theta=\theta-\beta \sum_{\mathcal{T}_q} \nabla_{\theta_i^{\prime}}{L}_{\mathit{ESD}}\left(Q_{source}^{(i)};\theta_{i}^{\prime}\right)$
\EndWhile
\end{algorithmic}
\label{algo:MAML4Span}
\end{algorithm}

\textbf{Entity Classification (EC)}

\setcounter{subsubsection}{0}
\subsubsection{EC Method}
For an entity $e_k=(x_f,...,x_{f+l})$, where $f$ is the first token index and $f+l$ is the last token index in the sequence, we employ the max-pooling to get its representation when $l>=1$ :
{
  \setlength{\abovedisplayskip}{3pt}
  \setlength{\belowdisplayskip}{2pt}
\begin{equation}
\hat{e}_k = max(h_f,...,h_{f+l})
\label{entrepmax}
\end{equation}}

Contrastive learning could enhance the consistency between entities within the same types and widen the distance between entities belonging to different types. As $e_k$ is the sample with certain type from the supervised $D_{source}$, we consider using the supervised contrastive learning to deepen the distinctiveness of entity types and augment representations of entities for entity classification improvement.

In details, given a batch with \textit{N} entities and the anchor index \textit{j} $\in$ \{1,2,...,\textit{N}\}, an entity-aware contrastive loss can be defined as follows: \par
{
  \setlength{\abovedisplayskip}{-5pt}
    \begin{scriptsize}{
\begin{equation}
    L_{CL} = \sum_{j} -\frac{1}{|P(j)|} \sum_{p \in P(j)} log \frac{exp(sim(z^j, z^p)/\tau)}{\sum_{a \in A(j)}exp(sim(z^j,z^a)/\tau)}
\end{equation}}\end{scriptsize}}where \textit{P(j)} is the positive set whose entities are from the same type with  $e_j$, $z^j$ is the result of transforming $\hat{e}_j$ through a projection network MLP, \textit{A(j)} is the set containing all \textit{N} entities except for $e_j$, sim(·,·) denotes the \textit{KL}-divergence for $K$-shot ($K>$1) but the squared euclidean distance for $K$-shot ($K$=1) referring to \cite{das2022container}, and $\tau$ is a temperature hyperparameter. 

After pulling entities of the same type together and entities of different types farther apart in the semantic space with contrastive learning, we discard the projection network at classification time following the previous works of contrastive learning \cite{chen2020simple,khosla2020supervisedcl,liupeipei2023msa}. 
Subsequently, prototypes for different types of entities are constructed based on the enhanced entity features. Let $\mathcal{S}_t$ denote the set of entities belonging to the $t$-th type in the support dataset $S_{source}^{(i)}$ of a task during training. The prototype for the $t$-th type is obtained by calculating the average of all entity features in this type.
\begin{equation}
    c_{t}(\mathcal{S})=\frac{1}{\left|S_{t}\right|} \sum_{{e}_m \in S_{t}} \hat{e}_m
\label{entproto}
\end{equation}

As a consequence, we can predict the type distribution of $e_k$ by calculating the distance between its entity representation $\hat{e}_k$ and all type prototypes:\par{\begin{small}{
\begin{equation}
            p_{soft}(e_k)=\frac{\exp \left(-d\left(c_{t}(\mathcal{S}), \hat{e}_k\right)\right)}{\sum_{ii=1}^{|\phi|} \exp \left(-d\left(c_{ii}(\mathcal{S}), \hat{e}_k\right)\right)}, t \in \{1,..,|\phi|\}
    \label{ecner}
\end{equation}}\end{small}}where $\phi$ is the entity type set and $|\phi|$=$N$, $d(\cdot, \cdot)$ is the distance function.
So, the entity classification loss is formalized as:
{
\begin{equation}
    L_{EC} = -\sum_{k=1}^{NK} y_klog(p_{soft}({e}_k))
    \label{enttypeloss}
\end{equation}}where $y_k$ is the truth type of ${e}_k$, $L_{\mathit{EC}}$ is the loss which we optimize.

\subsubsection{Training}
The training of the model in this stage (detailed in Algorithm \ref{algo:MAML4Proto}) adopts the MAML-ProtoNet approach proposed by Ma et al. \cite{matingting2022mamlprotype}, which combines the MAML method with prototype network techniques. Essentially, the key difference from MAML lies in treating the prototypes of entity types in the support set as type labels for entities of the same type during training.

In the inner-loop update, the model first uses Equation (\ref{ecner}) to predict the types for all true entity spans in the support set \( S_{source}^{(i)} \) of each task. Then, the entity classification loss is computed based on \( c^N_{k=1}(\mathcal{S}) \) and Equation (\ref{enttypeloss}). Finally, gradient computation is performed using Equation (\ref{onegradup}) to optimize the parameters.

In the outer-loop update, the training process is similar to entity span detection. The model first uses Equation (\ref{ecner}) to predict the types for all true entity spans in the query set \( Q_{source}^{(i)} \) of each task. Then, based on \( c^N_{k=1}(\mathcal{S}) \) and Equation (\ref{enttypeloss}), the entity classification loss is computed. Subsequently, as shown in Equation (\ref{mamllosssum}), the losses across all tasks are accumulated. Finally, gradient computation is performed using Equation (\ref{mamlgradupdate}) to optimize the parameters.

\begin{algorithm}
\caption{Training Process of Entity Classification Model Based on MAML-ProtoNet Algorithm}
\begin{algorithmic}[1] 
\Require $ D_{source}$: Training dataset for the source entity types
\Require $\alpha, \beta$: Learning rates for inner and outer-loop update
\State Randomly initialize $\theta$
\While{not done}
    \State Randomly select a batch of data $\mathcal{T}$, $\mathcal{T} \sim D_{source}$
    \ForAll{$\mathcal{T}$}
        \State Select a task $(S_{source}^{(i)}, Q_{source}^{(i)})$ from $\mathcal{T}$
        
        \State Obtain $c^N_{k=1}(\mathcal{S})$ by computing prototype vectors on $S_{source}^{(i)}$ using Equation (\ref{entproto})
        
        \State Based on $c^N_{k=1}(\mathcal{S})$ and the true entity spans in the support set $S_{source}^{(i)}$, compute $\nabla_{\theta} {L}_{\mathit{EC}}\left(S_{source}^{(i)}; {\theta}\right)$ using Equations (\ref{ecner}, \ref{enttypeloss})
        
        \State Perform gradient update: $\theta_{i}^{\prime} = \theta - \alpha \nabla_{\theta} {L}_{\mathit{EC}}\left(S_{source}^{(i)}; {\theta}\right)$
        
        \State Based on $c^N_{k=1}(\mathcal{S})$ and the true entity spans in the query set $Q_{source}^{(i)}$, compute $\nabla_{\theta_i^{\prime}}{L}_{\mathit{EC}}\left(Q_{source}^{(i)}; \theta_{i}^{\prime}\right)$ using Equations (\ref{ecner}, \ref{enttypeloss}) and save the results
    \EndFor
    \State Outer-loop update: $\theta = \theta - \beta \sum\nolimits_{\mathcal{T}_q} \nabla_{\theta_i^{\prime}}{L}_{\mathit{EC}}\left(Q_{source}^{(i)}; \theta_{i}^{\prime}\right)$
\EndWhile
\end{algorithmic}
\label{algo:MAML4Proto}
\end{algorithm}

\subsection{Finetuning Process}
Model finetuning aims to enable the model to adapt to new knowledge types during the domain transfer process. We finetune the trained models for entity detection and entity classification on the support dataset of the target domain $S_{target}$, using the same pattern as training process.

\subsection{Inference Process}
In the inference process, we test the finetuned model on query dataset $Q_{target}$, and we will present this stage by four parts: 

(1) We compute the representation $\hat{e}_k$ of each true entity in the support dataset $S_{target}$ by Eq.\ref{entrepmax} and build a \textit{key-value} datastore $\mathcal{D}_{knn}$ where \textit{key} is the entity representation and \textit{value} is entity type. 

(2) We input the query sentence $x'$ into the finetuned best model for entity span detection (ESD) to get all entities. Specifically, we apply the Viterbi algorithm \cite{forney1973viterbi} for decoding, and derive the entities from the maximum sequence probability of Eq.\ref{crfprob}:
{
  \setlength{\abovedisplayskip}{3pt}
  \setlength{\belowdisplayskip}{3pt}
\begin{equation}
    y^* = argmax_{y\in \mathbb{Y}}P(y|x')
\end{equation}}where $\mathbb{Y}$ is a set of all possible label sequences.

\begin{table*}[]
\centering
{
\caption{F1 scores with standard deviations(\%) on FewNERD. \dag denotes the results reported in \cite{matingting2022mamlprotype} and \cite{xue2024dbcp}. \ddag are the results we produce by LLM. * means the abbreviation of ‘contrastive learning’. The best results are in bold.}
\label{mainresultstable}
\begin{tabular}{lcccccccc}
\toprule 
\multicolumn{1}{c}{\multirow{3}{*}{\textbf{Models}}} & \multicolumn{4}{c}{\textbf{FewNERD-INTRA}}                                                                                                            & \multicolumn{4}{c}{\textbf{FewNERD-INTER}}                                                                                                            \\ \cmidrule(r){2-5}\cmidrule(r){6-9}
\multicolumn{1}{c}{}                        & \multicolumn{2}{c}{\textbf{1-2 shot}}                                     & \multicolumn{2}{c}{\textbf{5-10 shot}}                                    & \multicolumn{2}{c}{\textbf{1-2 shot}}                                     & \multicolumn{2}{c}{\textbf{5-10 shot}}                                    \\ \cmidrule(r){2-3}\cmidrule(r){4-5}\cmidrule(r){6-7} \cmidrule(r){8-9} 
\multicolumn{1}{c}{}                        & \multicolumn{1}{c}{5 way}      & \multicolumn{1}{c}{10 way}     & \multicolumn{1}{c}{5 way}      & 10 way                          & \multicolumn{1}{c}{5 way}      & \multicolumn{1}{c}{10 way}     & \multicolumn{1}{c}{5 way}      & 10 way                          \\ \cmidrule(r){1-1}\cmidrule(r){2-5}\cmidrule(r){6-9}
ProtoBERT\dag                                     & \multicolumn{1}{c}{23.45±0.92} & \multicolumn{1}{c}{19.76±0.59} & \multicolumn{1}{c}{41.93±0.55} & 34.61±0.59                      & \multicolumn{1}{c}{ 44.44±0.11} & \multicolumn{1}{c}{39.09±0.87} & \multicolumn{1}{c}{58.80±1.42} &  53.97±0.38                      \\ 
NNShot\dag                                       & \multicolumn{1}{c}{31.01±1.21} & \multicolumn{1}{c}{21.88±0.23}  & \multicolumn{1}{c}{35.74±2.36} & 27.67±1.06                      & \multicolumn{1}{c}{54.29±0.40} & \multicolumn{1}{c}{ 46.98±1.96} & \multicolumn{1}{c}{50.56±3.33} & 50.00±0.36                      \\ 
StructShot\dag                               & \multicolumn{1}{c}{35.92±0.69} & \multicolumn{1}{c}{25.38±0.84} & \multicolumn{1}{c}{38.83±1.72} & 26.39±2.59                      & \multicolumn{1}{c}{57.33±0.53} & \multicolumn{1}{c}{49.46±0.53} & \multicolumn{1}{c}{ 57.16±2.09}  & 49.39±1.77                      \\ 
 CONTAINER\dag     & \multicolumn{1}{c}{40.436}  & \multicolumn{1}{c}{ 33.84} & \multicolumn{1}{c}{53.70}  & \multicolumn{1}{c}{47.49}  & \multicolumn{1}{c}{ 55.95} & \multicolumn{1}{c}{ 48.35} & \multicolumn{1}{c}{ 61.83}  & \multicolumn{1}{c}{ 57.12} \\

ESD\dag                                           & \multicolumn{1}{l}{41.44±1.16}  & \multicolumn{1}{l}{32.29±1.10} & \multicolumn{1}{l}{ 50.68±0.94}  & \multicolumn{1}{l}{42.92±0.75}  & \multicolumn{1}{l}{66.46±0.49} & \multicolumn{1}{l}{59.95±0.69} & \multicolumn{1}{l}{74.14±0.80}  & \multicolumn{1}{l}{67.91±1.41} \\

MAML-ProtoNet\dag                                           & \multicolumn{1}{c}{ 52.04±0.44}  & \multicolumn{1}{c}{43.50±0.59} & \multicolumn{1}{c}{63.23±0.45}  & \multicolumn{1}{c}{56.84±0.14}  & \multicolumn{1}{c}{ 68.77±0.24} & \multicolumn{1}{c}{ 63.26±0.40} & \multicolumn{1}{c}{ 71.62±0.16}  & \multicolumn{1}{c}{68.32±0.10} \\

BDCP\dag  & \multicolumn{1}{c}{ 52.63}  & \multicolumn{1}{c}{-} & \multicolumn{1}{c}{-}  & \multicolumn{1}{c}{57.46}  & \multicolumn{1}{c}{ 69.59} & \multicolumn{1}{c}{ -} & \multicolumn{1}{c}{ -}  & \multicolumn{1}{c}{68.87} \\

ChatGPT\ddag                                           & 
\multicolumn{1}{c}{ \textbf{61.86}}  & \multicolumn{1}{c}{\textbf{55.61}} & 
\multicolumn{1}{c}{64.17}  & \multicolumn{1}{c}{54.79}  & 
\multicolumn{1}{c}{ 61.05} & \multicolumn{1}{c}{ 57.94} & 
\multicolumn{1}{c}{ 64.34}  & \multicolumn{1}{c}{58.58} \\
\hline
\textbf{MsFNER} (Ours)      & \multicolumn{1}{l}{{54.25±0.34}} & \multicolumn{1}{l}{{46.69±0.5}} & \multicolumn{1}{l}{\textbf{66.57±0.29}} & \multicolumn{1}{l}{\textbf{58.70±1.39}} & \multicolumn{1}{l}{\textbf{72.91±0.34}} & \multicolumn{1}{l}{\textbf{66.34±0.65}} & \multicolumn{1}{l}{\textbf{78.41±0.30}} & \multicolumn{1}{l}{\textbf{72.06±0.11}} \\ 
w/o cl*                      & \multicolumn{1}{c}{—}          & \multicolumn{1}{c}{—}          & \multicolumn{1}{c}{65.72±0.33} & 57.33±0.08 & \multicolumn{1}{c}{—}           & \multicolumn{1}{c}{—}          & \multicolumn{1}{c}{77.84±0.42} & 71.23±0.47          \\
w/o \textit{KNN}                                       & \multicolumn{1}{c}{53.86±0.47} & \multicolumn{1}{c}{46.24±0.29} & \multicolumn{1}{c}{66.04±0.65} & 57.56±0.41 & \multicolumn{1}{c}{72.75±0.29} & \multicolumn{1}{c}{65.38±0.72} & \multicolumn{1}{c}{77.99±0.25} & 71.92±0.14          \\ \bottomrule
\end{tabular}
}
\vspace{-0.3cm}
\end{table*}

\begin{table*}[]
  \centering
  {
\caption{The comparison of inference time between \textbf{MsFNER} and ChatGPT.}
\label{time}
\begin{tabular}{ccccccccccccc}
\toprule 
\multicolumn{1}{c}{\multirow{3}{*}{\textbf{Models}}} & \multicolumn{4}{c}{\textbf{FewNERD-INTRA}}                                                                                                            & \multicolumn{4}{c}{\textbf{FewNERD-INTER}} & \multicolumn{4}{c}{\textbf{FewAPTER}}                                                                                                           \\ \cmidrule(r){2-5}\cmidrule(r){6-9}\cmidrule(r){10-13}
\multicolumn{1}{c}{}  & \multicolumn{2}{c}{\textbf{1-2 shot}}  & \multicolumn{2}{c}{\textbf{5-10 shot}}                                    & \multicolumn{2}{c}{\textbf{1-2 shot}}   & \multicolumn{2}{c}{\textbf{5-10 shot}}
& \multicolumn{2}{c}{\textbf{1 shot}}   & \multicolumn{2}{c}{\textbf{3 shot}}
\\ \cmidrule(r){2-3}\cmidrule(r){4-5}\cmidrule(r){6-7} \cmidrule(r){8-9}\cmidrule(r){10-11} \cmidrule(r){12-13} 
\multicolumn{1}{c}{}                        & \multicolumn{1}{c}{5 way}      & \multicolumn{1}{c}{10 way}     & \multicolumn{1}{c}{5 way}      & 10 way                          & \multicolumn{1}{c}{5 way}      & \multicolumn{1}{c}{10 way}     & \multicolumn{1}{c}{5 way}      & 10 way& \multicolumn{1}{c}{4 way}      & \multicolumn{1}{c}{6 way}     & \multicolumn{1}{c}{4 way}      & 6 way                          \\ \cmidrule(r){1-1}\cmidrule(r){2-5}\cmidrule(r){6-9}\cmidrule(r){10-13}
\multicolumn{1}{l}{MsFNER}  & \multicolumn{1}{c}{0.3442} & \multicolumn{1}{c}{0.3808} & \multicolumn{1}{c}{0.5616} & 1.0546 & \multicolumn{1}{c}{0.3482} & \multicolumn{1}{c}{0.3806} & \multicolumn{1}{c}{0.6708} & 1.2266 &  0.5126 &  0.4981 &  0.7530 &  0.6055 \\ 
\multicolumn{1}{l}{ChatGPT} & \multicolumn{1}{c}{4.5581}  & \multicolumn{1}{c}{5.4897} & \multicolumn{1}{c}{8.1842}   & 9.3481  & \multicolumn{1}{c}{3.2633}  & \multicolumn{1}{c}{8.0579} & \multicolumn{1}{c}{4.7755} & 5.3409 & 2.3078 & 2.8610& 4.9470&  7.2926   \\ \hline
\end{tabular}}
\vspace{-0.4cm}
\end{table*}

(3) Firstly, we compute the representation $\hat{e}'_k$ of each detected entity ${e}'_k$ by Eq.\ref{entrepmax}. Then, we input $\hat{e}'_k$ into two independent modules: on one hand, we feed the $\hat{e}'_k$ into \textit{KNN} to obtain the predicted result $p_{knn}$ based on the $\mathcal{D}_{knn}$ \cite{wangshuhe2022knnner}, and on the other hand we pass the $\hat{e}'_k$ into the finetuned best model for entity classification to produce its predicted type distribution $p_{soft}$ by Eq.\ref{ecner}.

(4) The final prediction type of the detected entity ${e}'_k$ is jointly determined by \textit{KNN} and the entity classification model:
{
  \setlength{\abovedisplayskip}{4pt}
  \setlength{\belowdisplayskip}{4pt}
\begin{equation}
    p(y|e'_k) = \lambda p_{knn}(y|e'_k) + (1-\lambda) p_{soft}(y|e'_k)
\end{equation}}where the $\lambda$ is a hyper-parameter that makes a balance between \textit{KNN} and prototype classifier.

\begin{figure}[!ht]
  \centering
  \includegraphics[height=1.83in,width=2.3in]{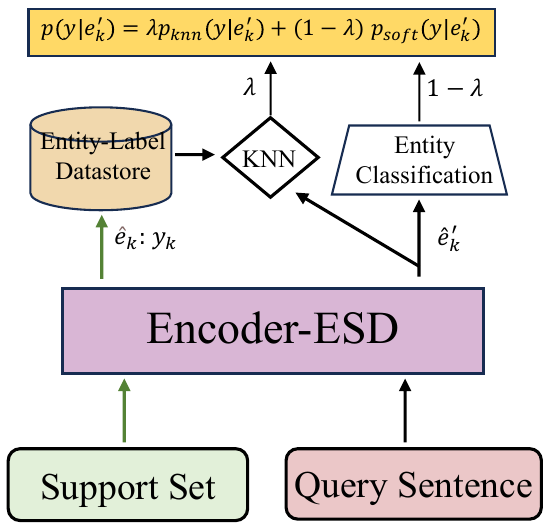}
  \caption{The schematic diagram of the inference process for entity classification.}  
  \label{inferencestage}
  \vspace{-0.3cm}
\end{figure}

\section{Experiments}
\label{sec:experiments}

\subsection{Experiments Setup}
We conduct experiments on the Tesla V100 GPU.
\subsubsection{Datasets}
\label{fewnerdataset}
We conduct experiments on the FewNERD dataset \cite{ding2021fewshotnerd} which is annotated with 8 coarse-grained and 66 fine-grained entity types. We evaluate our method on the two settings of FewNERD-INTRA and FewNERD-INTER, which are with the different grained types.
Furthermore, we follow the settings of ESD \cite{wangpeiyi2022esd} and adopt $N$-way $K\sim2K$-shot sampling method to construct tasks.
Another dataset we use for testing our method is FewAPTER, which is built by us based on the cybersecurity threat intelligence dataset APTER\cite{wang2020apter}.

Specifically, APTER is proposed by Wang et al.\cite{wang2020apter} within general data distributions, and it predefines 37 entity types under domain expert knowledge. In this paper, we refine and enhance the original APTER dataset, consolidating its entity types into 21 categories, resulting in a total of 28,250 entities. Inspired by the design methodology of the FewNERD dataset, we follow the knowledge transfer principle (from original entity type data $\rightarrow$ to new entity type data) and construct the FewAPTER dataset based on APTER for few-shot entity recognition research in the cybersecurity domain. Considering the data volume of each entity type, the 21 entity types are categorized into three subsets: the training set with 7 types \{VULID, FILE, SECTEAM, LOC, TOOL, APT, ACT\}, the validation set with 7 types \{SHA1, OS, URL, MAL, DOM, PROT, TIME\}, and the test set with 7 types \{ENCR, VULNAME, IP, MD5, SHA2, IDTY, EMAIL\}.

While constructing the dataset, it is necessary that the entity types in the training, validation, and test sets do not overlap. Therefore, a maximum of 7-way can be set. However, due to the imbalanced data distribution and the possibility of multiple entity types occurring in a single sentence, setting the dataset to 7-way limits the diversity of knowledge learned, making it challenging to adapt to and generalize for new tasks. 
To address this, this paper constructs four $N$-way $K$-shot datasets: \{4-way 1-shot, 4-way 3-shot, 6-way 1-shot, 6-way 3-shot\}. Each $N$-way $K$-shot dataset includes 4000 training samples, along with 800 validation samples and 800 test samples. During the construction of each $N$-way $K$-shot dataset (for training, validation, and testing), each task, denoted as $S^{(i)}$ or $Q^{(i)}$, is randomly selected from the corresponding dataset, consisting of $N$ entity types and $K$ examples. For example, when constructing a task $S^{(i)}_{source}$ or $Q^{(i)}_{source}$ for the 6-way 3-shot training set, 6 entity types are randomly chosen from the 7 available types, with 3 entity examples selected for each entity type.



\subsubsection{Parameter Settings} 
\label{paramsset}
We use grid search for hyperparameter settings and we train our model for 1,000 steps and choose the best model on validation dataset for testing. We use the pretrained language model uncased BERT-base as our encoder. In the entity span detection module, we adopt the BIOES tags. We use the AdamW optimizer with a learning rate of 3e-5 and 0.01 linear warmup steps. The $K$ in \textit{KNN} is set to 10 and the coefficient $\lambda$ of \textit{KNN} is 0.1 when adding \textit{KNN}. The batch size is set to 32, the max sequence length is set to 128 and we use a dropout probability of 0.2.

\subsection{Baselines}
\label{baselines}
We compare MsFNER with popular baselines as follows:
\begin{itemize}
    \item \textbf{ProtoBERT} \cite{fritzler2019few} employs the prediction of query labels by leveraging the similarity between the BERT hidden states derived from the support set and the query tokens.
    \item \textbf{CONTAINER} \cite{das2022container} employs BERT as the encoder and designs a token-level contrastive method to enhance the token embedding for few-shot NER.
    \item \textbf{NNShot} \cite{yangyi2020simplefewshotner} bears resemblance to ProtoBERT in its methodology, as it conducts predictions utilizing nearest neighbor algorithms.
    \item \textbf{StructShot} \cite{yangyi2020simplefewshotner} incorporates an augmented Viterbi decoder into the inference process atop NNShot's framework.
    \item \textbf{ESD} \cite{wangpeiyi2022esd} formulates few-shot sequence labeling as a span-level matching problem and decomposes it into span-related procedures.
    \item \textbf{MAML-ProtoNet} \cite{matingting2022mamlprotype} proposes a meta-learning approach with fusing MAML and prototype network for few-shot NER, which facilitates rapid adaptation to novel entity types. 
    \item \textbf{BDCP} \cite{xue2024dbcp} is a few-shot NER method with boundary discrimination and correlation purification, which explores the robustness of the model's cross domain transfer learning ability in textual adversarial attack scenarios.
    \item \textbf{ChatGPT} \cite{chatgpt} is an advanced large language model (LLM) developed by OpenAI, which is trained on diverse corpora of text and could generate human-like text.
\end{itemize}

\subsection{ChatGPT Prompt Template}
\label{chatgptpromptl}
In this section, we introduce the ChatGPT few-shot NER prompt template that we use. The prompt is composed of 3 parts: task description, few-shot cases, and input queries. 

In the task description, we initially elucidate the few-shot NER task required from ChatGPT. Subsequently, we provide ChatGPT with explicit instructions for implementing this task, including delineation of the entity types involved. To facilitate easier interpretation of model output, we utilize a structured output JSON format, a domain in which ChatGPT exhibits proficiency. The template is as follows:

\textit{Your task is to perform few-shot Named Entity Recognition. You could perform the task by the following actions:
\begin{enumerate}
    \item Do named entity recognition task to recognize the entities. Entity type is defined as \texttt{\$ent\_list}
    \item Check if the entity type is in the definition list.
    \item Output a JSON object that contains the following keys: text, entity\_list. Note that each item in entity\_list is a dictionary with “entity” and “type” as keys.
\end{enumerate}
}

\textit{Here are the given few-shot cases:}  \verb|```$case_input```|

\textit{Output: \$case\_output}

\textit{Recognize the entities in the text delimited by triple backticks.:} \verb|```$input_text```|

In the template above, the \$ent\_list denotes the list of entity types corresponding to $N$-way. The \$case\_input comprises the sentences in the support set corresponding to $K$-shot, while the \$case\_output represents the labels in the support set. Since ChatGPT is better at structured output, we use JSON format for output, with each output format being:
\textit{\{“entity”: entity text ; “type”: entity type\}}. The output will be merged into a list. The \$input\_text denotes the query data. When we input this prompt to ChatGPT, it will return a list of relevant entity-type dictionaries.

\subsection{Main results}
Table \ref{mainresultstable} illustrates main results of different methods on FewNERD dataset and Table \ref{fewnermainresults3} shows results on the FewAPTER dataset. These results demonstrate that \textbf{MsFNER} significantly outperforms previous few-shot NER approaches. We have several findings: 1) \textbf{MsFNER} attains average F1 improvements of 2.65\% and 4.44\% on FewNERD-INTRA and FewNERD-INTER, respectively, compared to the previous best few-shot NER method, MAML-ProtoNet. Furthermore, the greater improvement is on the FewAPTER dataset, which achieves an average F1 score improvement of 11.42\%. 
2) Although \textbf{MsFNER} outperforms the ChatGPT on the FewNERD dataset, it is slightly inferior to the LLM on FewAPTER. 
This is due to the characteristics of the FewAPTER dataset, which enable the model to exhibit strong learning capabilities within a specific domain, but with slightly reduced generalization ability for cross-domain learning.
3) Regardless of whether it is ChatGPT or small models, increasing the number of $K$-shot yields superior performance, while an increase in the number of $N$-way results in worse performance. This aligns with general human intuition: given a fixed number of entity types, having more examples leads to better learning accuracy; conversely, with a fixed number of examples, increasing the number of entity types makes learning more challenging.
4) From the performance evaluation of these results, it can be seen that the performance of LLM is quite stable, which significantly indicates that its influence from data and domain changes is relatively limited. This is primarily attributed to the large number of parameters in the model, which, unlike few-shot learning models, exhibits lower sensitivity to the number of training samples.

\subsection{Ablation Study}
To investigate the contributions of different modules of \textbf{MsFNER}, we conduct the ablation study by removing each of them at a time.
The results are shown at the bottom of Table \ref{mainresultstable} and Table \ref{fewnermainresults3}. 
Firstly, we remove the contrastive learning module. In the 1-shot experiment of this setting, \textit{KNN} is unnecessary as there is only one sample in the support dataset.
We can see that the removal of contrastive learning results in an average decrease of \textbf{0.905\%} on FewNERD and \textbf{0.745\%} on FewAPTER, which indicates the significant impact and necessity of contrastive learning for enhancing representation. 
Secondly, we remove the \textit{KNN} module and reserve the contrastive learning module. 
The removal of \textit{KNN} drops average \textbf{0.524\%} on FewNERD and \textbf{0.635\%} on FewAPTER. 
We can see that the impact of contrastive learning and \textit{KNN} changes with the number of $N$-way and $K$-shot. Removing either of them will lead to a decrease in the performance of \textbf{MsFNER}.

\subsection{Model Efficiency}
\label{modelefficiency}
To comprehensively evaluate \textbf{MsFNER} and ChatGPT, we compute their average inference time(s) in different settings, and the results are shown in Table \ref{time}. From the table, we can see that \textbf{MsFNER} is much faster than its LLM peer ChatGPT. According to the Occam's Razor, we think that \textbf{MsFNER} is a better choice for few-shot NER in application.

\begin{table}[]
  \centering
        \fontsize{7.2pt}{9pt}\selectfont
    \setlength{\tabcolsep}{4pt}
    \renewcommand{\arraystretch}{1.2}
\caption{\centering\enspace {F1 scores with standard deviations(\%) on FewAPTER}}
\begin{tabular}{lcccc}
\toprule 
\multicolumn{1}{c}{\multirow{3}{*}{\textbf{Models}}}  & \multicolumn{4}{c}{\textbf{FewAPTER}}     \\ \cmidrule(r){2-5}
\multicolumn{1}{c}{}  & \multicolumn{2}{c}{\textbf{1 shot}}  & \multicolumn{2}{c}{\textbf{3 shot}} \\ \cmidrule(r){2-3}\cmidrule(r){4-5}
\multicolumn{1}{c}{}  & \multicolumn{1}{c}{4 way}      & \multicolumn{1}{c}{6 way}     & \multicolumn{1}{c}{4 way}      & 6 way                             \\ \cmidrule(r){1-1}\cmidrule(r){2-5}

ESD\dag    & \multicolumn{1}{l}{33.83±0.57} & \multicolumn{1}{l}{29.97±0.42} & \multicolumn{1}{l}{41.60±0.27}  & \multicolumn{1}{l}{39.08±0.72} \\
 CONTAINER\dag      & \multicolumn{1}{c}{ 29.82} & \multicolumn{1}{c}{ 31.17} & \multicolumn{1}{c}{51.71}  & \multicolumn{1}{c}{ 50.65} \\ 
BDCP\dag  & \multicolumn{1}{c}{ 44.65±0.97} & \multicolumn{1}{c}{36.55±1.02} & \multicolumn{1}{c}{51.43±0.63} &  39.55±1.74                \\ 
MAML-ProtoNet\dag           & \multicolumn{1}{c}{{48.41±0.31}} & \multicolumn{1}{c}{ 39.32±0.19} & \multicolumn{1}{c}{ 49.51±0.47}  & \multicolumn{1}{c}{37.16±0.11} \\
ChatGPT\ddag & \multicolumn{1}{c}{ \textbf{59.14}   } & \multicolumn{1}{c}{ \textbf{57.65}   } & \multicolumn{1}{c}{ 61.14   }  & \multicolumn{1}{c}{\textbf{57.68}   } \\
\hline
\textbf{MsFNER}   & \multicolumn{1}{l}{{57.55±0.39}} & \multicolumn{1}{l}{{49.90±0.90}} & \multicolumn{1}{l}{\textbf{61.36±0.59}} & \multicolumn{1}{l}{{51.26±1.38}} \\ 
w/o cl* & \multicolumn{1}{c}{—}    & \multicolumn{1}{c}{—}  & \multicolumn{1}{c}{60.78±0.30} & 50.35±0.24         \\
w/o \textit{KNN}   & \multicolumn{1}{c}{56.47±0.69} & \multicolumn{1}{c}{49.23±0.80} & \multicolumn{1}{c}{61.21±0.17} & 50.62±1.23          \\ \bottomrule
\end{tabular}      
\label{fewnermainresults3}
\vspace{-0.1cm}
\end{table}

\subsection{Training and Validation Process}  
\begin{figure}[!ht]
  \centering
  \includegraphics[height=3.0in,width=3.4in]{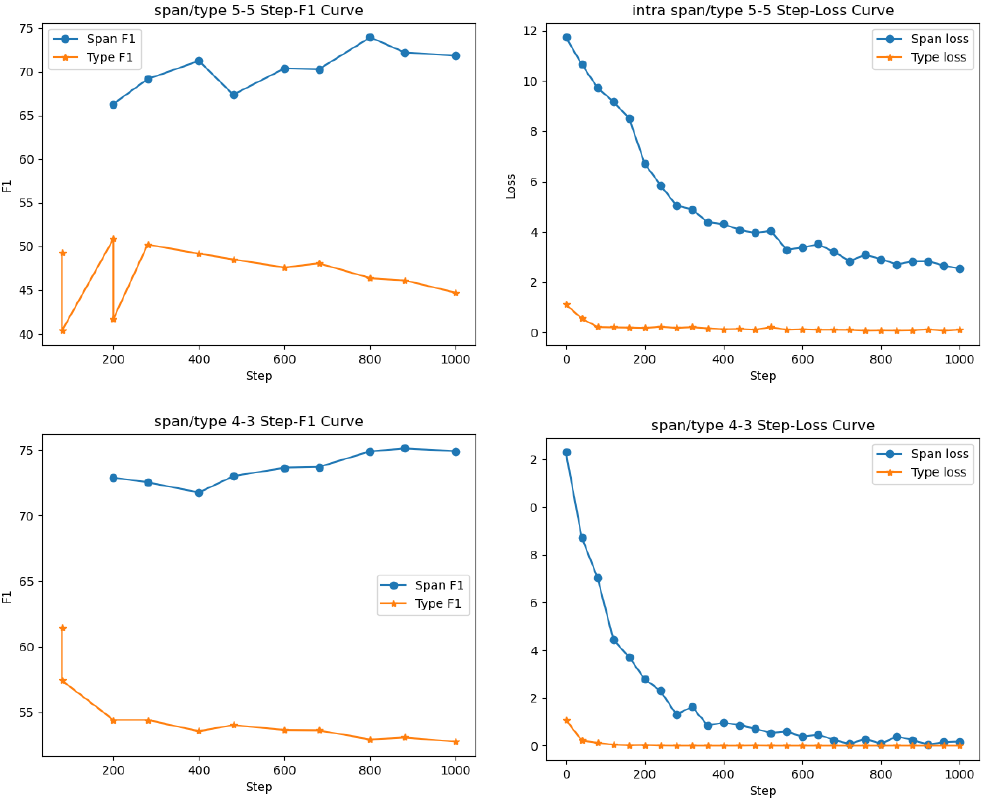}
  \caption{ The performance of training and validation stage on two different settings}  
  \label{trainvalid}
  \vspace{-0.3cm}
\end{figure}

As shown in Figure \ref{trainvalid}, this section presents the training and validation processes of \textbf{MsFNER} on the FEW-NERD dataset with a 5-way 5$\sim$10-shot setting (top two plots) and on the FEW-APTER dataset with a 4-way 3-shot setting (bottom two plots). Since the training dataset originates from the same source domain, the training loss decreases steadily over the training epochs (right two plots). However, during fine-tuning on the validation set, fluctuations occur due to cross-domain knowledge transfer (left two plots). As training progresses, slight overfitting appears on the validation dataset due to the limited number of supervised samples.

\section{Conclusion}
\label{sec:Conclusion}

In this paper, we propose \textbf{MsFNER}, a novel approach to addressing the challenges of a large number of negative samples and computational overhead in the domain of few-shot named entity recognition. \textbf{MsFNER} is implemented in three stages: training, fine-tuning, and inference. During the training and fine-tuning stages, we first detect the entity spans and subsequently recognize the entity types. To enhance the learning of entity representations, we introduce a contrastive learning module. In the inference stage, we replace the contrastive learning module with a \textit{KNN} module to further improve inference performance. The final inference output is determined by combining the \textit{KNN} results and the type classification predictions. We conduct experiments on the FewNERD and FewATPER datasets to evaluate the performance of \textbf{MsFNER}. Additionally, we compare our approach with ChatGPT-3.5 on few-shot tasks, and the results demonstrate that \textbf{MsFNER} can achieve performance comparable to ChatGPT-3.5 to some extent.

\bibliographystyle{IEEEbib}
\bibliography{refs}

\begin{thebibliography}{10}

\bibitem{yadav2018nersurvey}
Vikas Yadav and Steven Bethard,
\newblock ``A survey on recent advances in named entity recognition from deep learning models,''
\newblock in {\em Proceedings of the 27th International Conference on Computational Linguistics}, Santa Fe, New Mexico, USA, Aug. 2018, pp. 2145--2158.

\bibitem{lijing2022nersurvey}
Jing Li, Aixin Sun, Jianglei Han, and Chenliang Li,
\newblock ``A survey on deep learning for named entity recognition,''
\newblock {\em IEEE Transactions on Knowledge and Data Engineering}, vol. 34, no. 1, pp. 50--70, 2022.

\bibitem{lample2016neuralner}
Guillaume Lample, Miguel Ballesteros, and et~al.,
\newblock ``Neural architectures for named entity recognition,''
\newblock in {\em Proceedings of the 2016 Conference of the North {A}merican Chapter of the Association for Computational Linguistics: Human Language Technologies}, San Diego, California, June 2016, pp. 260--270.

\bibitem{mahovy2016neuralner}
Xuezhe Ma and Eduard Hovy,
\newblock ``End-to-end sequence labeling via bi-directional {LSTM}-{CNN}s-{CRF},''
\newblock in {\em Proceedings of the 54th Annual Meeting of the Association for Computational Linguistics (Volume 1: Long Papers)}, Berlin, Germany, Aug. 2016, pp. 1064--1074.

\bibitem{fritzler2019few}
Alexander Fritzler, Varvara Logacheva, and Maksim Kretov,
\newblock ``Few-shot classification in named entity recognition task,''
\newblock in {\em Proceedings of the 34th ACM/SIGAPP Symposium on Applied Computing}, 2019, pp. 993--1000.

\bibitem{Snell2017tokenfewner1}
Jake Snell, Kevin Swersky, and Richard Zemel,
\newblock ``Prototypical networks for few-shot learning,''
\newblock in {\em Advances in Neural Information Processing Systems}, 2017, vol.~30.

\bibitem{hou2020fewshotnerthransferlearning}
Yutai Hou, Wanxiang Che, Yongkui Lai, and et~al.,
\newblock ``Few-shot slot tagging with collapsed dependency transfer and label-enhanced task-adaptive projection network,''
\newblock in {\em Proceedings of the 58th Annual Meeting of the Association for Computational Linguistics}, Online, July 2020, pp. 1381--1393.

\bibitem{yangyi2020simplefewshotner}
Yi~Yang and Arzoo Katiyar,
\newblock ``Simple and effective few-shot named entity recognition with structured nearest neighbor learning,''
\newblock in {\em Proceedings of the 2020 Conference on Empirical Methods in Natural Language Processing (EMNLP)}, Online, Nov. 2020, pp. 6365--6375.

\bibitem{das2022container}
Sarkar Snigdha~Sarathi Das, Arzoo Katiyar, Rebecca Passonneau, and Rui Zhang,
\newblock ``{CONT}ai{NER}: Few-shot named entity recognition via contrastive learning,''
\newblock in {\em Proceedings of the 60th Annual Meeting of the Association for Computational Linguistics (Volume 1: Long Papers)}, Dublin, Ireland, May 2022, pp. 6338--6353.

\bibitem{wangpeiyi2022esd}
Peiyi Wang, Runxin Xu, Tianyu Liu, and et~al.,
\newblock ``An enhanced span-based decomposition method for few-shot sequence labeling,''
\newblock in {\em Proceedings of the 2022 Conference of the North American Chapter of the Association for Computational Linguistics: Human Language Technologies}, Seattle, United States, July 2022, pp. 5012--5024.

\bibitem{yudian2021fewshotner}
Dian Yu, Luheng He, Yuan Zhang, and et~al.,
\newblock ``Few-shot intent classification and slot filling with retrieved examples,''
\newblock in {\em Proceedings of the 2021 Conference of the North American Chapter of the Association for Computational Linguistics: Human Language Technologies}, Online, June 2021, pp. 734--749.

\bibitem{matingting2022mamlprotype}
Tingting Ma, Huiqiang Jiang, Qianhui Wu, and et~al.,
\newblock ``Decomposed meta-learning for few-shot named entity recognition,''
\newblock in {\em Findings of the Association for Computational Linguistics: ACL 2022}, Dublin, Ireland, May 2022, pp. 1584--1596.

\bibitem{finn2017maml}
Chelsea Finn, Pieter Abbeel, and Sergey Levine,
\newblock ``Model-agnostic meta-learning for fast adaptation of deep networks,''
\newblock in {\em Proceedings of the 34th International Conference on Machine Learning - Volume 70}. 2017, ICML'17, p. 1126–1135, JMLR.org.

\bibitem{chen2020simple}
Ting Chen, Simon Kornblith, Mohammad Norouzi, and Geoffrey Hinton,
\newblock ``A simple framework for contrastive learning of visual representations,''
\newblock {\em arXiv preprint arXiv:2002.05709}, 2020.

\bibitem{khosla2020supervisedcl}
Prannay Khosla, Piotr Teterwak, Chen Wang, Aaron Sarna, Yonglong Tian, Phillip Isola, Aaron Maschinot, Ce~Liu, and Dilip Krishnan,
\newblock ``Supervised contrastive learning,''
\newblock {\em arXiv preprint arXiv:2004.11362}, 2020.

\bibitem{liupeipei2023msa}
Peipei Liu, Xin Zheng, Hong Li, and et~al.,
\newblock ``Improving the modality representation with multi-view contrastive learning for multimodal sentiment analysis,''
\newblock in {\em ICASSP 2023 - 2023 IEEE International Conference on Acoustics, Speech and Signal Processing (ICASSP)}, 2023, pp. 1--5.

\bibitem{forney1973viterbi}
G.D. Forney,
\newblock ``The viterbi algorithm,''
\newblock {\em Proceedings of the IEEE}, vol. 61, no. 3, pp. 268--278, 1973.

\bibitem{xue2024dbcp}
Xiaojun Xue, Chunxia Zhang, Tianxiang Xu, and Zhendong Niu,
\newblock ``Robust few-shot named entity recognition with boundary discrimination and correlation purification,''
\newblock in {\em AAAI}, 2024.

\bibitem{wangshuhe2022knnner}
Shuhe Wang, Xiaoya Li, Yuxian Meng, and et~al.,
\newblock ``$ k $ nn-ner: Named entity recognition with nearest neighbor search,''
\newblock {\em arXiv preprint arXiv:2203.17103}, 2022.

\bibitem{ding2021fewshotnerd}
Ning Ding, Guangwei Xu, Yulin Chen, and et~al.,
\newblock ``Few-{NERD}: A few-shot named entity recognition dataset,''
\newblock in {\em Proceedings of the 59th Annual Meeting of the Association for Computational Linguistics and the 11th International Joint Conference on Natural Language Processing (Volume 1: Long Papers)}, Online, Aug. 2021, pp. 3198--3213.

\bibitem{wang2020apter}
Xuren Wang, Mengbo Xiong, Yali Luo, and et~al.,
\newblock ``Joint learning for document-level threat intelligence relation extraction and coreference resolution based on gcn,''
\newblock in {\em 2020 IEEE 19th International Conference on Trust, Security and Privacy in Computing and Communications (TrustCom)}, 2020, pp. 584--591.

\bibitem{chatgpt}
chatgpt,
\newblock ,'' \url{https://openai.com/blog/chatgpt}, 2023.

\end{thebibliography}

\end{document}